\documentclass[letterpaper, 10 pt, conference]{ieeeconf}  % Comment this line out if you need a4paper

\IEEEoverridecommandlockouts                              % This command is only needed if 
\usepackage{graphics} % for pdf, bitmapped graphics files
\usepackage{epsfig} % for postscript graphics files
\usepackage{mathptmx} % assumes new font selection scheme installed
\usepackage{times} % assumes new font selection scheme installed
\usepackage{amsmath} % assumes amsmath package installed
\usepackage{amssymb}  % assumes amsmath package installed
\usepackage{xcolor}
\usepackage{multirow} %用于画图
\usepackage{dsfont}
\usepackage[T1]{fontenc} %解决LaTeX Error: Command \DH unavailable in encoding OT1.
\usepackage{url} %可能的引用链接
\usepackage{booktabs} % for images
\usepackage{subcaption} % for images
\usepackage{cite} % save some space for continuous citation numbers

\title{\LARGE \bf
TopoLiDM: Topology-Aware LiDAR Diffusion Models for Interpretable and Realistic LiDAR Point Cloud Generation 
}

\author{Jiuming Liu$^{\dagger1}$, Zheng Huang$^{\dagger1}$, Mengmeng Liu$^{2}$, Tianchen Deng$^{1}$,\\ Francesco Nex$^{2}$, Hao Cheng$^{2}$, and Hesheng Wang$^{*1}$% <-this % stops a space
\thanks{$\dagger$The first two authors contribute equally.}
\thanks{*This work was supported in part by the Natural Science Foundation of China under Grant 62225309, U24A20278, 62361166632, U21A20480 and 62403311. Corresponding Author: Hesheng Wang ({\tt\small wanghesheng@sjtu.edu.cn}).}% <-this % stops a space
\thanks{$^{1}$ School of Automation and Intelligent Sensing, Shanghai Jiao Tong University, Shanghai 200240. Key Laboratory of System Control and Information Processing, Ministry of Education of China, Shanghai 200240. ({\tt\small liujiuming@sjtu.edu.cn})}%
\thanks{$^{2}$ University of Twente, Netherlands.
        {(\tt\small m.liu-1@utwente.nl)}}%
}

\begin{document}

\maketitle
\thispagestyle{empty}
\pagestyle{empty}

%%%%%%%%%%%%%%%%%%%%%%%%%%%%%%%%%%%%%%%%%%%%%%%%%%%%%%%%%%%%%%%%%%%%%%%%%%%%%%%%
\begin{abstract}

LiDAR scene generation is critical for mitigating real-world LiDAR data collection costs and enhancing the robustness of downstream perception tasks in autonomous driving. However, existing methods commonly struggle to capture geometric realism and global topological consistency. Recent LiDAR Diffusion Models (LiDMs) predominantly embed LiDAR points into the latent space for improved generation efficiency, which limits their interpretable ability to model detailed geometric structures and preserve global topological consistency. To address these challenges, we propose TopoLiDM, a novel framework that integrates graph neural networks (GNNs) with diffusion models under topological regularization for high-fidelity LiDAR generation. Our approach first trains a topological-preserving VAE to extract latent graph representations by graph construction and multiple graph convolutional layers. Then we freeze the VAE and generate novel latent topological graphs through the latent diffusion models. We also introduce 0-dimensional persistent homology (PH) constraints, ensuring the generated LiDAR scenes adhere to real-world global topological structures.  Extensive experiments on the KITTI-360 dataset demonstrate TopoLiDM’s superiority over state-of-the-art methods, achieving improvements of 22.6\% lower Fréchet Range Image Distance (FRID) and 9.2\% lower Minimum Matching Distance (MMD). Notably, our model also enables fast generation speed with an average inference time of 1.68 samples/s, showcasing its scalability for real-world applications. We will release the related codes at https://github.com/IRMVLab/TopoLiDM.

\end{abstract}
%leverages a latent diffusion paradigm to encode range images into a compact latent space, enabling efficient two-stage training while maintaining structural coherence. Crucially, we By harmonizing diffusion-based generation with topological graph learning, TopoLiDM advances the frontier of realistic LiDAR simulation for safety-critical autonomous systems.

%%%%%%%%%%%%%%%%%%%%%%%%%%%%%%%%%%%%%%%%%%%%%%%%%%%%%%%%%%%%%%%%%%%%%%%%%%%%%%%%
\section{INTRODUCTION} 

\begin{figure}[t]
\centering
\includegraphics[width=0.9\linewidth]{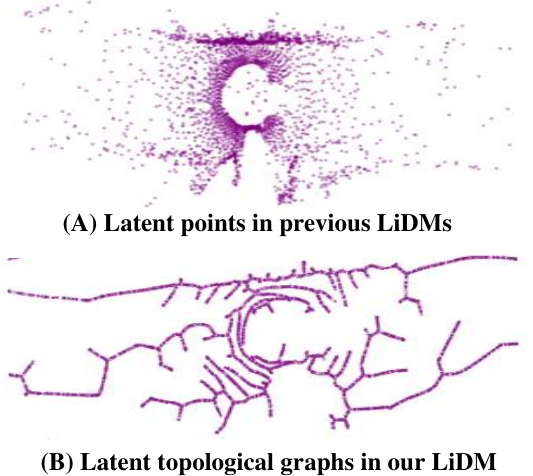}
\caption{Comparison with previous LiDAR Diffusion Models (LiDMs). Previous LiDMs commonly adopt latent points as the variables in diffusion models, which fail to consider the global topological consistency. Our method treats topologically-regularized graphs as latent variables to generate interpretable LiDAR graphs with geometric correlations.}
\label{fig:compare}
%\vspace{-2pt}
\end{figure}

Recently, LiDAR data has achieved remarkable research focus with the advantages of accurate geometric measurements \cite{liu2024point,yan2025turboregturbocliquerobustefficient,liu2025mamba4d}, facilitating various downstream tasks in autonomous driving \cite{li2020deep,deng2025mne,liu2024dvlo}, robotic localization \cite{liu2023translo,deng2024compact}, and SLAM system \cite{zhu2024semgauss,zhu2024sni,deng2024plgslam}. However, the LiDAR data collection process remains difficult due to the substantial costs of physical LiDAR sensors and performance degradation in corner cases, e.g., extreme weather conditions \cite{hu2024rangeldm}. Therefore, many researchers resort to deep generative models for realistic and low-cost LiDAR scene generation \cite{caccia2019deep,zyrianov2022learning,xiong2023learning,nakashima2024lidar,ran2024towards}.

%is a crucial component in enhancing the robustness and reliability of perception systems for autonomous driving. High-fidelity LiDAR data can significantly improve the performance of downstream tasks such as object detection, segmentation, and scene understanding, ultimately leading to safer and more efficient autonomous navigation.

However, existing methods for LiDAR data generation often face significant challenges in maintaining geometric fidelity, topological consistency, and interpretable ability. Caccia et al. \cite{caccia2019deep} propose an unsupervised method with Generative Adversarial Networks (GAN) \cite{goodfellow2020generative} and Variational Autoencoders (VAEs) \cite{kingma2013auto}. Nonetheless, the training process of GAN suffers from instability and mode collapse, and the generation samples commonly lack geometric details \cite{ran2024towards}. More recently,  many researchers also explore LiDAR generation based on diffusion models \cite{zyrianov2022learning,nakashima2024lidar,ran2024towards}. These methods predominantly embed LiDAR points into the latent space to improve the generation efficiency as in Fig. \ref{fig:compare} (A), but this limits the interpretable ability to model geometric correlations and preserve global topological consistency. 
% diffusion models \cite{ho2020denoising,nichol2021improved} have witnessed remarkable advances in the generation quality and diversity, which achieve tremendous success in image generation \cite{rombach2022high} and object-level point cloud generation \cite{luo2021diffusion}.
%This limitation is particularly obvious when dealing with complex scenes where objects exhibit non-rectangular or non-curve patterns \emph{maybe need some figures here?}, leading to the loss of critical structural details during the generation process.

% Moreover, the adoption of range images as the primary data representation in existing LiDMs further exacerbates the problem. While range images provide a convenient and efficient way to represent LiDAR data, they fail to capture the intrinsic topological properties of the underlying 3D scenes. This shortcoming is especially pronounced when generating LiDAR data for autonomous driving scenarios, where the global topology of the environment plays a vital role in navigation and decision-making. 

To address these challenges, we propose TopoLiDM, a novel framework that integrates graph neural networks (GNNs) with diffusion models under topological regularization for interpretable and high-fidelity LiDAR point generation. TopoLiDM consists of a two-stage training paradigm. First, a topology-preserving VAE learns how to compress input LiDAR range images into compact topological graphs and integrates 0-dimensional persistent homology (PH) constraints \cite{kumar2024glidr} into the learning objective, which ensures that synthesized LiDAR scenes maintain the intrinsic global topological properties inherent to real-world distributions. The VAE encoder incorporates a computational graph transformation module that converts input points into graph-structured representations, followed by a series of GNN layers to establish topological relationships. Then, a diffusion-based generation module is designed based on compact and interpretable topological graphs as the intermediate latent variables, effectively preserving geometric correlations. This compact representation not only optimizes the generation efficiency but also enhances the model's capacity to capture long-range dependencies and enforce consistency in global topological structures. Finally, generated topological graphs are decoded into 3D LiDAR points by the pre-trained topology-aware decoder of VAE. 

%Combining the advantages of GNNs and diffusion models, our method achieves interpretable high-fidelity LiDAR generation as well as comparative generation speed compared to the baseline methods.

Overall, our key contributions are as follows:
\begin{itemize}
    \item We propose TopoLiDM, a novel framework integrating GNNs with diffusion models under topological regularization. With compact topological graphs as latent representations, our method can realize fast, interpretable, and high-fidelity LiDAR generation.
    \item We design a topology-aware VAE module that captures long-range dependencies and enforces 0-dimensional persistent homology constraints, ensuring the generated LiDAR scenes adhere to the global topology rules of real-world environments.
    \item Extensive experiments on the KITTI-360 dataset demonstrate that TopoLiDM achieves significant improvements over state-of-the-art methods. Notably, our model surpasses baseline methods by 22.6\% in Fréchet Range Image Distance (FRID) and 9.2\% in Minimum Matching Distance (MMD) with fast inference time, showcasing great potential for practical applications.    
\end{itemize}

\begin{figure*}[t]
\centering
\includegraphics[width=0.95\linewidth]{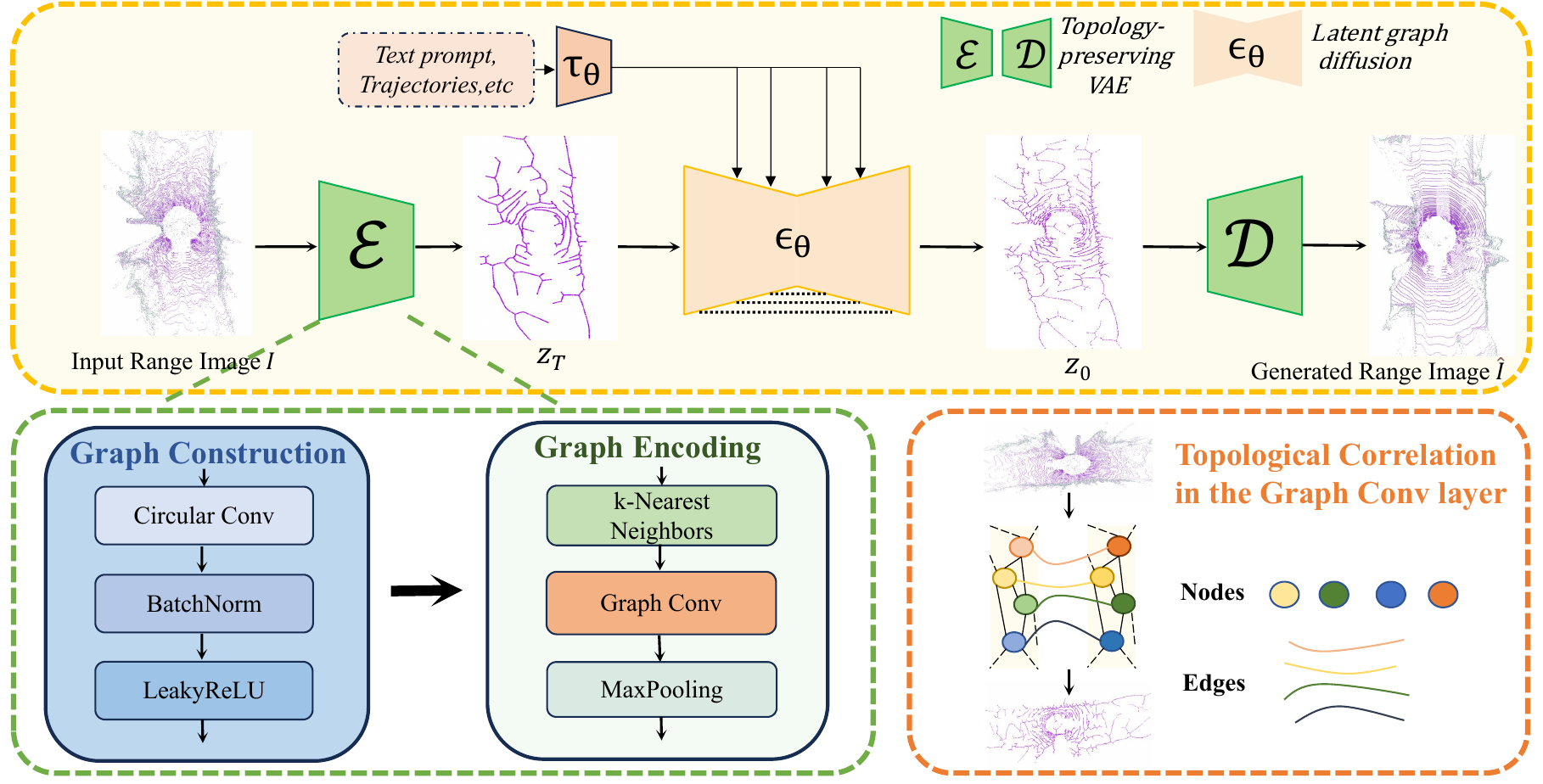}
%\vspace{-5pt}
\caption{The overall architecture of TopoLiDM consists of two cascaded stages: (1) \textbf{Topology-Preserving Autoencoding} that compresses LiDAR points into graph-structured latent representations. (2) \textbf{Diffusion-based Topological LiDAR Generation} that leverages the low-dimensional, semantically-rich representation for a more efficient and interpretable denoising process.}
\label{fig:pipe}
%\vspace{-10pt}
\end{figure*}

\section{RELATED WORK}

\subsection{Deep Generative Models}
Deep generative models, including GAN \cite{goodfellow2020generative}, VAE \cite{kingma2013auto}, etc., have revolutionized the field of computer vision and machine learning, enabling the synthesis of high-fidelity 2D images \cite{rombach2022high} and 3D points \cite{luo2021diffusion}. Recently, diffusion models \cite{ho2020denoising} have emerged as a more advanced generative backbone with better generation quality and diversity, leveraging a progressive denoising process to generate realistic samples. Latent diffusion models (LDMs) \cite{rombach2022high} further improve computational efficiency by operating in a perceptually equivalent latent space. Diffusion models have been successfully applied to various tasks, including text-to-image synthesis \cite{ruiz2023dreambooth}, image-to-image translation \cite{saharia2022palette}, and 3D shape generation \cite{luo2021diffusion}. However, the exploration of diffusion models in LiDAR data generation remains limited, primarily due to intrinsic challenges from sparse and irregular LiDAR points.

\subsection{LiDAR Data Generation}
LiDAR data generation is a critical task for robot localization \cite{yan2024ml,yan2025hemora,deng2025mcn} and autonomous driving\cite{liu2023regformer,liu2024difflow3d} because it enables the creation of training data to enhance the performance in downstream perception tasks \cite{yang2024visual}. Traditional methods for LiDAR data generation often rely on physics-based simulators or data-driven approaches \cite{dosovitskiy2017carla}. While these methods can produce realistic LiDAR scenes, they often fail to capture the intricate details and global topological properties of real-world environments. Recent works have explored the use of deep generative models for LiDAR data generation, including variants of GANs \cite{caccia2019deep}, VAEs \cite{xiong2023learning}, and diffusion models \cite{nakashima2024lidar,ran2024towards}. However, these methods often struggle to maintain geometric fidelity, topological consistency, and interpretable ability in latent space. 

%For instance, existing LiDMs\cite{ran2024towards} based on CNNs and range images often fail to model irregular geometric patterns and preserve global topological structures, leading to suboptimal performance in downstream tasks.

\subsection{Topological Learning for 3D Points}
Topological machine learning \cite{gabrielsson2020topology} is an emerging field that combines topological data analysis with machine learning techniques to capture and quantify the global shape and evolution of topological features in datasets. Persistent homology (PH) \cite{chen2019topological} is a key tool in this field, enabling the extraction of topological invariants such as connected components, loops, and voids. Recent works including \cite{chen2019topological, clough2020topological, liu2016applying} have explored the integration of PH with deep learning models, demonstrating its potential for improving the robustness and generalization of neural networks. In the context of LiDAR data generation, topological regularization has been shown to enhance the realism and fidelity of generated scenes by enforcing global topological constraints \cite{kumar2024glidr, zhou2022learning}. 
This motivates us to design topological regularization and PH-based constraints in diffusion models to ensure the generated LiDAR scenes adhere to the intrinsic global topology of real-world environments.

%%%%% 2.20
\section{Method}
%TopoLiDM designs a topology-aware latent diffusion paradigm for realistic LiDAR generation as in Fig. \ref{fig:pipe}, combining the high efficiency of latent space modeling with accurate graph-based topological learning. 

%The architecture operates in two cascaded stages: (1) Topology-Preserving Autoencoding that compresses LiDAR points into graph-structured latent representations. (2) Diffusion-based Topological LiDAR Generation that leverages the low-dimensional, geometrically-rich representation for a more efficient and interpretable generation process.

\subsection{Overall Architecture}  
As in Fig. \ref{fig:pipe}, given an input LiDAR range image \(I \in \mathbb{R}^{H \times W \times C}\), TopoLiDM first encodes it into a compact graph latent feature \(Z = \mathcal{E}(I) \in \mathbb{R}^{h \times w \times c}\) using a graph encoder \(\mathcal{E}\). The encoded graph $Z$ consists of nodes \(V\) representing local point clusters and edges \(E\) that encode spatial relationships through hierarchical graph convolutions. 

In the diffusion stage, a UNet \cite{ronneberger2015u} backbone \(\epsilon_\theta\) learns to denoise latent representations \(Z_t\) over \(T\) timesteps, guided by diverse input conditions, e.g., text, trajectory, etc. The diffusion process operates entirely in the latent space, enabling efficient LiDAR generation.  
%$L2$ constraints from original practice

Finally, the decoder \(\mathcal{D}\) reconstructs the range image \(\hat{I} = \mathcal{D}(Z)\) utilizing a combination of upsampling modules and circular convolutional blocks \cite{ran2024towards}.

\subsection{Topology-Preserving Autoencoding}  
To achieve efficient generation on latent space and preserve the accurate topology relationships, we first employ a topology-preserving VAE inspired by \cite{kumar2024glidr} to compress the 
% high-dimensional inputs $x\in \mathbb{R}^{H \times W \times 1}$ 
input LiDAR range image \(I \in \mathbb{R}^{H \times W \times C}\)
into a compact latent representation $\mathbf{z_0} \in \mathbb{R}^{h \times w \times c}$, where the downsampling factors $f_v=\frac{H}{h}$, $f_h=\frac{W}{w}$ determine the vertical and horizontal compression ratio, respectively. This latent representation $\mathbf{z_0}$ captures the essential topological features by constructing hierarchical graph layers while significantly reducing computational costs.

\textbf{Graph Construction:} Following \cite{han2022vision}, we divide an input range image $I$ into $N$ patches, then apply circular convolution layers to embed each patch into a feature vector $x_i \in \mathbb{R}^D$, where $D$ is the feature dimension and $i = 1, 2, ..., N$. These features can be viewed as a set of unordered nodes $X = [x_1,  x_2,...,x_N]$. We also add a positional encoding vector to each node in order to represent the positional information:

\[
x_{i} \leftarrow x_i + e_i,
\]  
%\vspace{-6pt}

where $e_i \in \mathbb{R}^D$.
Edges are established by correlating each node to its $k$-nearest neighbors. %%%%%% def of V, E ??? \(G = (V, E)\)

% convert the range image \(I\) into a graph \(G\) by treating each valid pixel as a node. Node features \(v_i \in V\) encode the pixel's depth, intensity, and normalized coordinates \((a_i, b_i)\). Edges  connect each node to its \(k\)-nearest neighbors in the azimuth-elevation plane, with edge weights computed as:  where \(\sigma\) controls the receptive field.  

\textbf{Hierarchical Graph Encoding:} The encoder \(\mathcal{E}\) then employs multiple graph layers for effective graph encoding. Specifically, each layer employs message passing to gather information from both immediate and distant neighbors, followed by a max-pooling operation to update the node embeddings as \cite{kumar2024glidr}:

\begin{align*}
% h_v^{(l)} &= \text{LeakyReLU}\left( W_{\text{conv}}^{(l)} \cdot \sum_{u \in N_v} h_u^{(l-1)} \right), \\
h_v^{(l)} &= \max_{u \in N_v} \left\{ \text{LeakyReLU}\left( W_{\text{conv}}^{(l)} \cdot \sum_{u \in N_v} h_u^{(l-1)} \right) \right\},
\end{align*}
where $N_v$ is $k$ nearest neighbors for each node $v$, \(W_{\text{conv}}^{(l)}\) is the learnable weight matrix for layer \(l\), and $h_u^{(l)}$ is the embedding of node $u$ at layer $l$.

\begin{figure*}[t]
\centering
\includegraphics[width=0.9\linewidth]{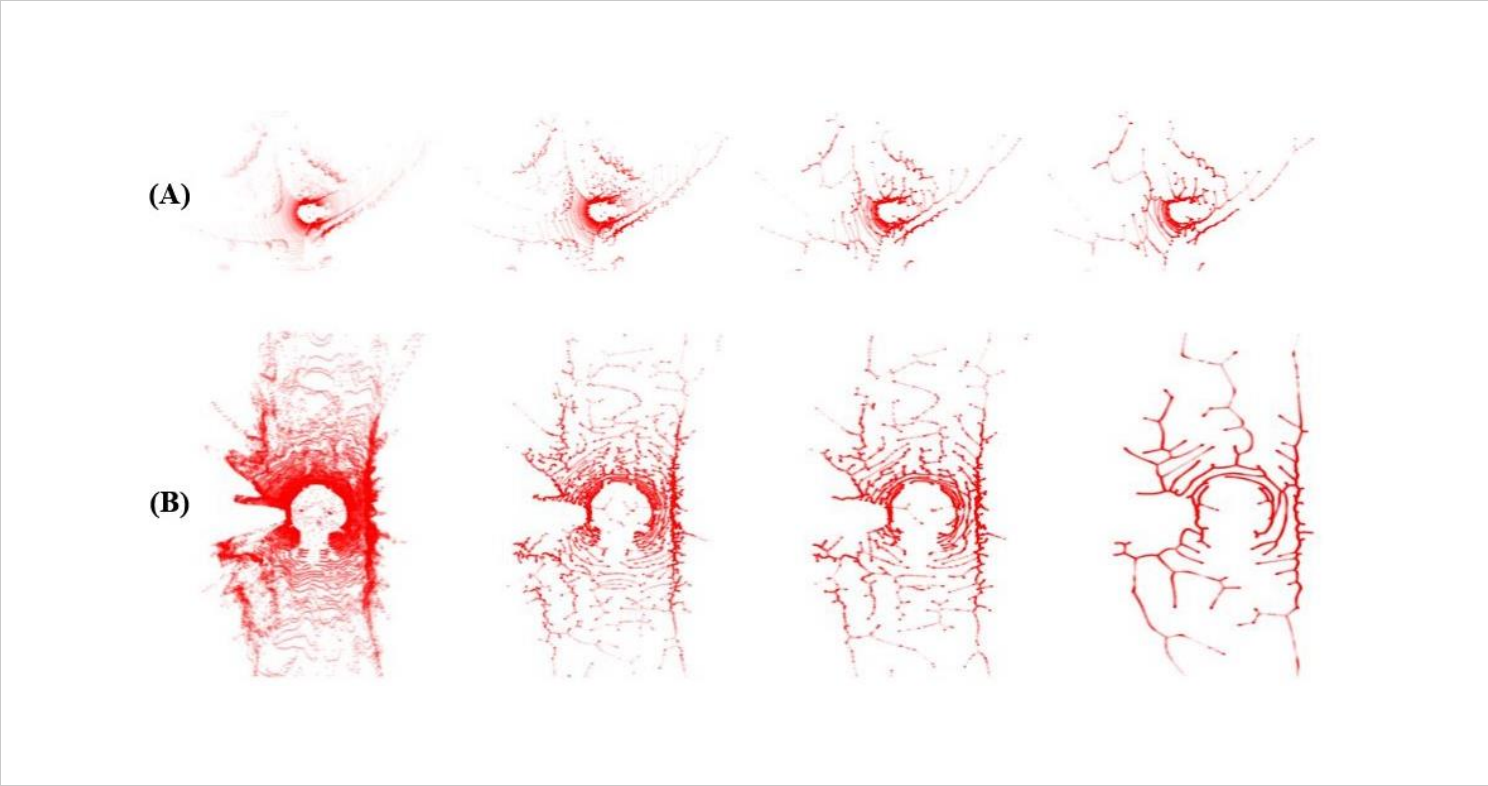}
%\vspace{-2pt}
\caption{Visualization of the progressive graph encoding process for latent topological graph creation. From the left to the right, compact graphs are progressively constructed with more topological correlations. The last columns indicate the final latent topological graphs which would be used as the intermediate latent variables in diffusion models.}
\label{fig:topology}
%\vspace{-6pt}
\end{figure*}

This process effectively projects computational graph's node features into higher-dimensional feature $\mathbf{z_0}$, thereby enriching their representation capacity for high-level semantic information. Upon completion of each layer, the original graph structure is discarded, and the updated node embeddings serve as the input for the subsequent layer. This recursive transformation enables the aggregation of long-range topological relationships, progressively encoding global topological features as in Fig. \ref{fig:topology}. The encoder is constructed by stacking $L$ LiDAR graph layers, progressively enhancing the global topological features of the input data. 
%As layers progress, nodes that were initially non-adjacent in the original LiDAR graph may become neighbors in the higher-dimensional embedding space.

\textbf{LiDAR Decoding:} Then the encoded graph representation (as in the last column in Fig. \ref{fig:topology}) will serve as the variable in diffusion models (Section \ref{sec:diffusion}) for the iterative LiDAR denoising and generation. After obtaining generated latent graphs, we leverage a circular-convolutional decoder \(\mathcal{D}\) \cite{ran2024towards} to recover LiDAR range images from the latent graph space.

\textbf{Persistent Homology Constraints:} To enforce the alignment between generated LiDAR graphs and the real-world LiDAR points, we enforce a 0-dimensional PH constraints on the decoded range image \(\hat{I}\) and penalize deviations from expected topological structures as in Section \ref{sec:loss}.

\subsection{Diffusion-based Topological LiDAR Generation}
\label{sec:diffusion}
In the second stage, we freeze the pre-trained VAE and utilize a latent diffusion model (LDM) \cite{rombach2022high} to efficiently generate latent topological features $\mathbf{z}_0$. 

During the forward diffusion process, noise is progressively added to the ground truth latent variable \( \mathbf{z}_0 \) over a series of time steps \( t \in \{1, 2, \dots, T\} \), resulting in a noisy latent variable \( \mathbf{z}_t \). This process can be formally described as:
%\vspace{-6pt}
\begin{equation*} %%%%% 这里编号省略
\mathbf{z}_t = \sqrt{1 - \beta_t}   \mathbf{z}_{t-1} + \sqrt{\beta_t} \epsilon_t, \quad \epsilon_t \sim N(0, \mathbf{I}),
\end{equation*}

where \( \beta_t \) is a predefined noise schedule that controls the variance of the noise added at each step.

During the backward denoising process, we employ a time-conditional U-Net network \( \epsilon_{\theta}(\mathbf{z}_t, t) \) to predict the noise \( \epsilon_t \) added to the noisy latent variable \( \mathbf{z}_t \). We leverage the standard loss in diffusion models as introduced in the next section to encourage the accurate prediction of the noise added at each time step, thereby enabling the reverse diffusion process to generate diverse and high-quality latent topological variables \( \mathbf{z}_0 \) from noisy latent variable \( \mathbf{z}_t \). Notably, our LiDAR generation can also be conditioned by additional guidance signals as in Fig. \ref{fig:pipe}, e.g., text prompts, trajectories, which offers more controllability and diversity for conditional LiDAR generation.

%The training process of TopoLiDM thus comprises two distinct phases: the dimensionality reduction via VAE and the distribution modeling via LDM. model the unconditional distribution \( p(\mathbf{z}_0) \), that 

\subsection{Training Process and Loss Functions}
\label{sec:loss}
Our training process consists of two parts: (1) First, we train the topology-preserving VAE, which effectively extracts the latent geometric graph features from the original input LiDAR points and the reverse decoding process. (2) Second, we freeze the parameters of encoder \( \mathcal{E} \) and decoder \( \mathcal{D} \) in the VAE, and continue to train the latent diffusion models.

% \textbf{Stage 1: VAE Optimization:} The loss functions in VAE combines a 0-dimensional persistent homology (PH) loss for effective latent topological feature learning, and also a reconstruction loss. To enforce alignment with real-world LiDAR topology, we compute a 0-dimensional persistent homology (PH) loss on the decoded range image \(\hat{I}\) and penalize deviations from the expected connected component structure. Specifically, we calculate the persistence diagram \(PD_0 = \{(b_i, d_i)\}\) for \(\hat{I}\), where \(b_i\) and \(d_i\) denote the birth/death of connected components. The topology loss encourages a single dominant component as:  %%%%%% given a filtration threshold \(\alpha\), 
% $$
% \mathcal{L}_{\text{topo}} = \sum_{i=2}^{N} \mathds{1} \{i \neq 1\}(b_i - d_i), % using the dsfont here
% $$  
\noindent\textbf{Stage 1: VAE Optimization:} The loss functions in VAE combine a 0-dimensional PH loss for effective latent topological feature learning, and also a reconstruction loss.  The former is a regularization term that leverages 0-dimensional PH to ensure the reconstructed point cloud follows the single connected component, which can be viewed as the global LiDAR topology graphs as in the last column of Fig. \ref{fig:topology}, thereby preserving the global topological structure of the real LiDAR scan. Formally, the topological loss is defined as:
$$
\mathcal{L}_{\text{topo}} = \sum_{i=2}^{M} \mathds{1} \{i \neq 1\}(b_i - d_i), % using the dsfont here
$$  
% where $M$ is the number of connected components, and $(b_i-d_i)$ measures the persistence of the $i$-th component. 
where $M$ is the number of connected components, \(b_i\) and \(d_i\) denote the \textit{birth} and \textit{death} times of the \(i\)-th connected component (0-dimensional topological feature) in the persistence diagram computed via sublevel set filtration over the LiDAR point cloud \cite{kumar2024glidr}.  % To stay short and neat,  maybe hint readers to refer to the paper of GLiDR?
The difference \((b_i - d_i)\) quantifies the \textit{persistence} (lifetime) of the component. 
% Minimizing this loss encourages the model to produce point clouds that adhere to the global topological shape of the LiDAR scan. 
We also leverage the PH-based loss at the computational graphs from the second and fourth graph layer's output in the encoder $\mathcal{E}$ to further improve topological embedding. The loss function in the VAE \( \mathcal{L}_{VAE} \) is formulated as:
\[
\mathcal{L}_{\text{VAE}} = \underbrace{\|I - \hat{I}\|_1}_{\text{Reconstruction}} + \underbrace{
\mathcal{L}_{\text{topo}}(\hat{I})+
\mathcal{L}_{\text{topo}}(L_2)+
\mathcal{L}_{\text{topo}}(L_{4})
}
_{\text{Topology}},
\]
 where \( L_{2} \) and \( L_{4} \) represent the updated node embeddings of the computational graphs obtained from the second and fourth layers of the encoder, respectively. 
% To further enhance topological fidelity, we also apply the PH-based loss at the 4-dimensional and the 16-dimensional computational graph at the outputs of the second and fourth graph layer of the encoder $\mathcal{E}$. The 0-dimensional persistent homology (0-dim PH) based regularization applied to the second and fourth layers of the encoder ensures that the intermediate graph representations adhere to the same homological constraints that define the global LiDAR topology and are compliant with 0-dim PH. The single connected component-based global LiDAR backbone acts as a regularizer at these intermediate layers as well.
% minimizing the persistence of spurious components. This is differentiable via the soft persistence approximation from GLiDR [19].  

\noindent\textbf{Stage 2: Diffusion Training:} During the second training stage, we generate the denoised latent graphs from the latent diffusion model (LDM). To minimize the distribution differences between the ground truth noise and predicted noise, we formulate the standard diffusion loss as:
\begin{equation*}
\mathcal{L}_{\text{LDM}} = \mathbb{E}_{ \epsilon \sim N(0, \mathbf{I}),\, t \sim \text{Uniform}(1, T)} \left[ \left\| \epsilon - \epsilon_{\theta}(\mathbf{z}_t, t, \tau_\theta(y)) \right\|^2_2 \right],
\end{equation*}
%\mathbf{z}_0 \sim p_{\text{data}}(\mathbf{z}_0),y,\,
where \( y \) is the optional text condition embedded by \( \tau_\theta(y) \) as in Fig. \ref{fig:pipe} that guides the diffusion model.

\begin{figure*}[t]
\centering
\includegraphics[width=1\linewidth]{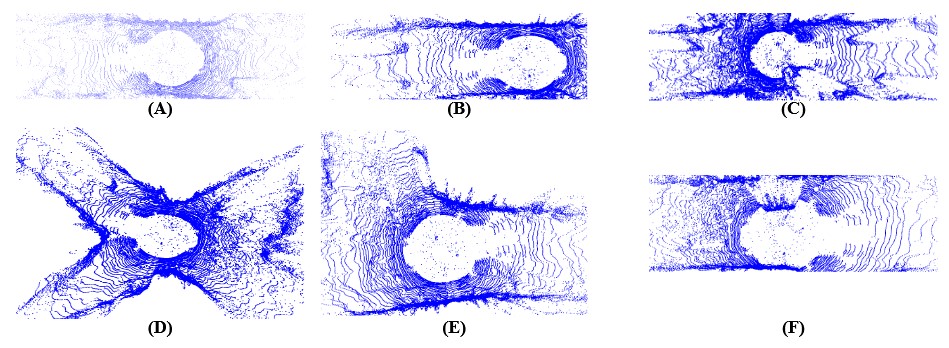}
\vspace{-10pt}
\caption{Visualization of the unconditional generation samples.}
\label{fig:uncon}
\vspace{-6pt}
\end{figure*}

\begin{table}[t]
\caption{Unconditional generation results on the KITTI-360 dataset \cite{liao2022kitti}. Best values are highlighted in \textbf{bold}.} %
\label{tab:performance comparison}
\vspace{-10pt}
\begin{center}
\resizebox{1.0\columnwidth}{!}
	{
\begin{tabular}{lccc|cc}
\hline
\multirow{2}{*}[-2pt]{Method} & \multicolumn{3}{c}{\textbf{Perceptual}} & \multicolumn{2}{|c}{\textbf{Statistical}}\\
& FRID$\downarrow$ & FSVD$\downarrow$ & FPVD$\downarrow$ & JSD$\downarrow$ & MMD$\downarrow$(10$^{-4}$) \\
\hline
Noise   & 3277     & 497.1    & 336.2    & 0.360     & 32.09    \\
LiDARGAN~\cite{caccia2019deep} & 1222     & 183.4    & 168.1    & 0.272    & 4.74     \\
LiDARVAE~\cite{caccia2019deep} & 199.1    & 129.9    & 105.8    & 0.237    & 7.07     \\
LiDARGen~\cite{zyrianov2022learning} (1160s) & 129.1 & 39.2 & 33.4 & 0.188 & 2.88 \\ 
\hline
% \rule{0pt}{0pt} % 添加额外的间距
LiDARGen~\cite{zyrianov2022learning} (50s) & 2051 & 480.6 & 400.7 & 0.506 & 9.91 \\
LDM(50s)~\cite{rombach2022high} & 199.5 & 70.7 & 61.9 & 0.236 & 5.06 \\
LiDM(50s)~\cite{ran2024towards}  & 158.8 & \bf53.7 & 42.7 & 0.213 & 4.46 \\
TopoLiDM (50s)  & \textbf{118.5} & 55.9 & \bf40.1 & \bf0.211 & \bf3.86 \\ % seed=750 
\hline
\end{tabular}}
\vspace{-6pt}
\end{center}
\end{table}

\begin{table}[t]
\caption{Efficiency analysis for LiDAR generation.}
\label{tab:inferspeed}
\vspace{-10pt}
\begin{center}
\resizebox{1.0\columnwidth}{!}
		{
\begin{tabular}{l |c |l|c}
\hline
Method & Speed (samples/s) $\uparrow$&Method & Speed (samples/s) $\uparrow$ \\
\hline
LiDARGen~\cite{zyrianov2022learning} &  0.08 & LDM~\cite{rombach2022high} & 1.31 \\
LiDM~\cite{ran2024towards} & 1.74 &Ours & 1.68 \\
\hline
\end{tabular}}
\end{center}
\vspace{-6pt}
\end{table}

%This training procedure effectively enables the LDM to learn the underlying distribution of latent topological features \( \mathbf{z}_0 \), which in turn boosts the generation quality of LiDAR points by sampling from this learned distribution.

\section{EXPERIMENTS}
\subsection{Experimental Setup}
\noindent\textbf{Datasets:} We evaluate the performance of TopoLiDM on the KITTI-360 dataset \cite{liao2022kitti}, which provides high-quality 64-beam LiDAR data for urban driving scenarios, making it ideal for studying distributions of outdoor LiDAR points and also assessing the performance of generated LiDAR scenes in terms of both realism and topological coherence.

\noindent\textbf{Metrics:} We evaluate the performance of TopoLiDM using metrics that assess both the geometric accuracy and the semantic consistency of the generated LiDAR scenes \cite{ran2024towards}:
(1) Fréchet Range Image Distance (FRID): This metric evaluates the perceptual similarity between the generated range images and the real range images.
(2) Fréchet Sparse Volume Distance (FSVD): This metric assesses how well the sparse volumes derived from the generated range images match those from the real data.
(3) Fréchet Point-based Volume Distance (FPVD): This metric compares the geometric accuracy between point clouds derived from the generated range images and the ground truth point clouds.
(4) Jensen-Shannon Divergence (JSD): This is a statistical measure of the distribution similarity between the generated and real LiDAR scenes.
(5) Minimum Matching Distance (MMD): This metric evaluates the minimum distance required to match each point in the generated scene with its counterpart in the real scene.
%, providing a measure of the visual realism achieved by our model , focusing on the geometric fidelity at a volumetric level , offering a direct measure of the geometric accuracy  , capturing the overall consistency and coherence  , providing a quantitative assessment of the geometric precision
%, This metric evaluates the perceptual similarity between the generated range images and the real range images.: This metric assesses how well the sparse volumes derived from the generated range images match those from the real data.providing a measure of the visual realism achieved by our model ,: This metric compares the geometric accuracy between point clouds derived from the generated range images and the ground truth point clouds.: This is a statistical measure of the distribution similarity between the generated and real LiDAR scenes. : This metric evaluates the minimum distance required to match each point in the generated scene with its counterpart in the real scene.focusing on the geometric fidelity at a volumetric level , offering a direct measure of the geometric accuracy  , capturing the overall consistency and coherence  , providing a quantitative assessment of the geometric precision

\noindent\textbf{Implementation Details:} We evaluate our TopoLiDM on a single NVIDIA RTX 4090 GPU. The model is trained using the Adam optimizer with $\beta_1 = 0.5$, $\beta_2 = 0.9$. 
The base learning rate is $4.5 \times 10^{-6}$ for the VAE module, $1.0 \times 10^{-6}$ for LDM, and both follow a cosine annealing schedule with a maximum period of 100 epochs, dynamically adjusting the learning rate until it reaches 0.
The dimension of input range images is set as \(H=64\), \(W=1024\), \(C=1\). For the Graph Construction module, we set the number of neighboring points as \(k=20\). The number of hierarchical graph layers is set as 4. The downsampling factors $f_v$ is set to 4, $f_h$ is set to 8. The latent dimension in the diffusion model is set to \(d=16\) for a compact representation. The diffusion process is configured with 1,000 timesteps, while the sampling procedure employs 50 steps for efficient generation.

\subsection{Unconditional Generation Results}
\noindent\textbf{Quantitative Comparison Results:} We compare TopoLiDM with several state-of-the-art methods for unconditional LiDAR data generation, such as LiDM~\cite{ran2024towards}: a recent SOTA approach focusing on generating realistic LiDAR scenes using diffusion models. We also compare other models including LiDARGAN~\cite{caccia2019deep}, LiDARVAE~\cite{caccia2019deep}, LiDARGen~\cite{zyrianov2022learning}, LDM~\cite{rombach2022high}. As shown in Table \ref{tab:performance comparison}, our TopoLiDM outperforms the chosen baseline method LiDM \cite{ran2024towards} in most metrics. TopoLiDM also outperforms other state-of-the-art methods in key metrics. It achieves the lowest FRID (118.5), FPVD (40.1), JSD (0.211), and MMD (3.86), indicating superior geometric accuracy and semantic consistency.

\begin{figure}[t]
    \centering
    \begin{subfigure}{0.45\textwidth}
        \includegraphics[scale=0.2]{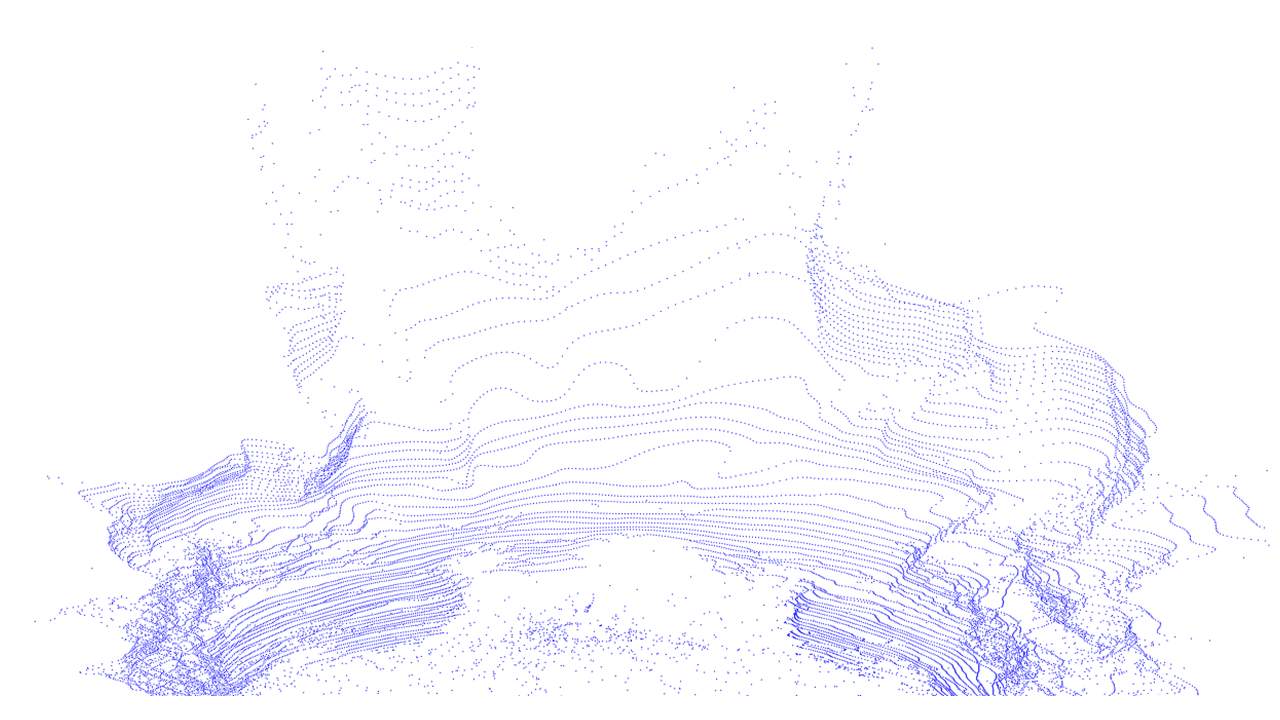}
        \caption{Input: \textit{An empty road without any object.}}
        \label{fig:empty_road}
    \end{subfigure}
    % \hfill % 添加一些水平空间
    \begin{subfigure}{0.45\textwidth}
        \includegraphics[scale=0.3]{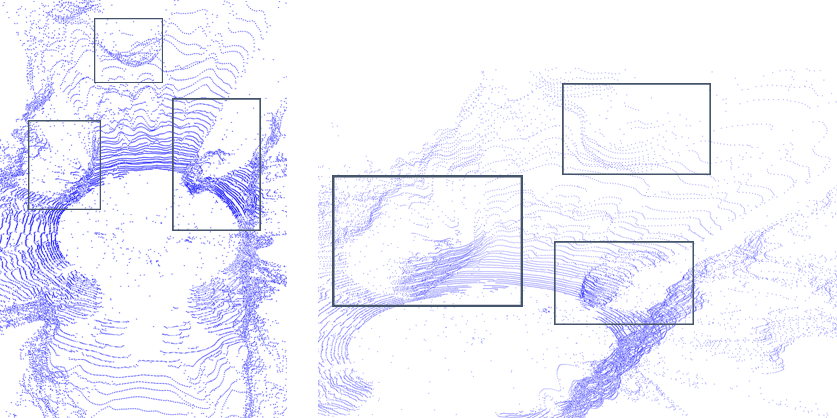}
        \caption{Input: \textit{With a few cars near the ego-car.}}
        \label{fig:cars}
    \end{subfigure}
    \caption{Visualization samples of the conditional text-to-LiDAR generation.}
    \label{fig:overall}
    \vspace{-6pt}
\end{figure}
%

% \begin{figure}[t]
%     \centering
%     % \hfill % 添加一些水平空间
   
%     \includegraphics[scale=0.4]{images/text2lidar/cars.png}
%         % \caption{Input: \textit{With a few cars near the ego-car.}}
%         % \label{fig:cars}
%     %\vspace{-2pt}
%     \caption{Visualization of the conditional text-to-LiDAR generation. (Input: \textit{With a few cars near the ego-car.})}
%     \label{fig:overall}
%     %\vspace{-4pt}
% \end{figure}

\begin{table}[t]
\centering
\caption{Ablation studies of Topology-aware Encoder.}
\label{tab:topoencoder}
\vspace{-6pt}
\resizebox{1.0\columnwidth}{!}
		{
\begin{tabular}{lccccc}
\toprule
Method & FRID $\downarrow$ & FSVD $\downarrow$ & FPVD $\downarrow$ & JSD $\downarrow$ & MMD $\downarrow$ $\times (10^{-4})$ \\
\midrule

w/o Topology-aware Encoder~\cite{ran2024towards} & 158.8 & \bf53.7 & 42.7 & 0.213 & 4.46 \\ 
with Topology-aware Encoder & \bf118.5 & 55.9 & \bf40.1 & \bf0.211 & \bf3.86 \\
%% here, borrow the data from lidm baseline, using the original convolutional encoder
\bottomrule
\end{tabular}}
%\vspace{-6pt}
\end{table}

\setlength{\tabcolsep}{5.0mm}
\begin{table}[t]
\centering
\caption{Ablation studies of the depth of GNN layers.}
\label{tab:gnn_depth}
\vspace{-6pt}
\resizebox{1.0\columnwidth}{!}
		{
\begin{tabular}{cccccc}
\toprule
Layers & FRID $\downarrow$ & FSVD $\downarrow$ & FPVD $\downarrow$ & JSD $\downarrow$ & MMD $\downarrow$ $\times (10^{-4})$ \\
\midrule
2 & 140.9 & 61.7 & 48.0 & 0.217 & 4.31 \\
3 & 133.2 & 65.7 & 44.1 & 0.219 & 4.09 \\
4 & \textbf{118.5} & \textbf{55.9} & \bf40.1 & \bf0.211 & \bf3.86 \\
\bottomrule
\vspace{-6pt}
\end{tabular}}
\end{table}

\noindent\textbf{Inference Speed:} We also evaluate the generation speed in Table \ref{tab:inferspeed}. TopoLiDM demonstrates fast generation speed with an average of 1.68 samples per second, offering potential for real-world applications. Compared with the chosen baseline LiDM \cite{ran2024towards} (1.74 samples per second), introducing our designed latent topologically-regularized diffusion models could boost the accuracy a lot with a slightly lower speed. 

\noindent\textbf{Qualitative Visualization:} In Fig. \ref{fig:uncon}, we visualize several generated LiDAR point cloud samples from our method TopoLiDM. The visualizations demonstrate the model's ability to produce realistic and topologically consistent LiDAR scenes, effectively capturing the topological structures of urban driving environments.

\subsection{Conditional LiDAR Generation}
\label{subsec:conLiD}
Additionally, we explored the potential capability of TopoLiDM to generate LiDAR point clouds conditioned by text descriptions. The visualization in Fig. \ref{fig:overall} shows that our model can produce realistic LiDAR scenes that align with the given text inputs, further highlighting its potential for creating training data for downstream fields.

\begin{table}[t]
\centering
\caption{Ablation studies of Topological Loss.}
\label{tab:topoloss}
\vspace{-6pt}
\resizebox{1.0\columnwidth}{!}{

\begin{tabular}{lccccc}
\toprule
Method & FRID $\downarrow$ & FSVD $\downarrow$ & FPVD $\downarrow$ & JSD $\downarrow$ & MMD $\downarrow$ $\times (10^{-4})$ \\
\midrule

w/o TopoLoss & 136.0 & \bf55.1 & 40.5 & 0.218 & 4.29 \\
with TopoLoss &\bf 118.5 & 55.9 & \bf40.1 & \bf0.211 & \bf3.86 \\
\bottomrule
\end{tabular}}
\vspace{-6pt}
\end{table}

\begin{table}[t]
\centering
\caption{Ablation studies of the of our latent diffusion model.}
\label{tab:ldm}
\vspace{-6pt}
\resizebox{1.0\columnwidth}{!}{
\begin{tabular}{lccccc}
\toprule
Method & FRID $\downarrow$ & FSVD $\downarrow$ & FPVD $\downarrow$ & JSD $\downarrow$ & MMD $\downarrow$ $\times (10^{-4})$ \\
\midrule

w/o LDM & 1991 & 437.1 & 261.7 & 0.351 & 12.43 \\
with LDM & \bf118.5 & \bf55.9 & \bf40.1 & \bf0.211 & \bf3.86 \\
\bottomrule
\end{tabular}}
%\vspace{-6pt}
\end{table}

\begin{table}[h]
\centering
\caption{Ablation studies of the number of sampling steps in our diffusion model.}
\label{tab:steps}
\vspace{-6pt}
\resizebox{1.0\columnwidth}{!}{
\begin{tabular}{cccccc}
\toprule
Steps & FRID $\downarrow$ & FSVD $\downarrow$ & FPVD $\downarrow$ & JSD $\downarrow$ & MMD $\downarrow$ $\times (10^{-4})$ \\
\midrule

25 & 133.1 & 60.7 & 43.5 & 0.215 & 4.29 \\
50 & 118.5 & 55.9 & 40.1 & \bf0.211 & 3.86 \\
100 & 114.5 & 54.7 & 39.8 & 0.214 & \bf3.77 \\
200 & \bf113.1 & \bf53.7 & \bf38.9 & 0.213 & 3.78 \\
\bottomrule
\end{tabular}}
\vspace{-6pt}
\end{table}

\subsection{Ablation Studies}

To analyze contributions of different components in TopoLiDM, we conduct ablation studies on KITTI-360.

\noindent\textbf{Topological Encoder in VAE:} We analyze the impact of using a topology-aware encoder versus a standard CNN encoder\cite{ran2024towards} without topological constraints. The results are presented in Table \ref{tab:topoencoder}. The model with the topology-aware encoder significantly outperforms the model without it, achieving lower FRID and MMD scores. This demonstrates the effectiveness of incorporating topological constraints in the encoder to preserve global topological consistency.

\noindent\textbf{Depth of GNN Layers:} We also investigate the impact of the depth of GNN layers in the encoder. The results are shown in Table \ref{tab:gnn_depth}. The proposed encoder with 4 layers achieves the best performance. Reducing the depth to 3 layers results in a noticeable drop in performance, particularly in FRID and FSVD, indicating that the model struggles to capture sufficient global topological context.

\noindent\textbf{Topological Loss Analysis:} We also evaluate the impact of the topological loss (TopoLoss). The results are shown in Table \ref{tab:topoloss}, where the model with the topological loss achieves better performance in terms of FRID and MMD, highlighting the importance of enforcing topological consistency.

\noindent\textbf{Diffusion Models:} We also conduct ablation studies for the designed latent diffusion models. As in Table \ref{tab:ldm}, we remove the diffusion models, only utilizing topology-aware VAE for the LiDAR generation, which has much worse performance. We also compare different sampling steps in diffusion models in Fig. \ref{tab:steps}. More sampling steps may slightly increase the generation accuracy but will incur much longer inference time. Thus, we choose the sampling step as 50 in our diffusion model for the trade-off between accuracy and efficiency.

\section{CONCLUSION}

We proposed TopoLiDM, a novel framework that integrates graph neural networks with diffusion models under topological regularization, significantly enhancing the fidelity and interpretable ability of LiDAR point cloud generation. Extensive experiments on the KITTI-360 dataset demonstrate the competitive performance of our TopoLiDM. For our future work, we would design more condition signals like trajectory, and also evaluate the proposed algorithm on additional datasets. 

%Our approach first trains a topological-preserving VAE to extract latent graph representations by a graph construction and multiple graph layers. Then we freeze the VAE and generate novel latent topology variables through the latent diffusion models. The proposed topology-aware graph diffusion module and multi-scale graph denoiser effectively capture long-range dependencies and preserve global topological structures. Ablation studies further highlight the contributions of the key components, including the graph construction method, GNN layer depth, and topological loss. Future work may explore advanced topological constraints and extend the framework to other applications, such as multi-sensor fusion and semantic segmentation, to further advance the field of autonomous driving and robotic perception. % or focus on the VAE's decoder part maybe refine the curve-like structure in real LiDAR data

\addtolength{\textheight}{-1cm}   % This command serves to balance the column lengths

%%%%%%%%%%%%%%%%%%%%%%%%%%%%%%%%%%%%%%%%%%%%%%%%%%%%%%%%%%%%%%%%%%%%%%%%%%%%%%%%

%%%%%%%%%%%%%%%%%%%%%%%%%%%%%%%%%%%%%%%%%%%%%%%%%%%%%%%%%%%%%%%%%%%%%%%%%%%%%%%%



% \section*{ACKNOWLEDGMENT}
% \textit{pass}

% The preferred spelling of the word ÒacknowledgmentÓ in America is without an ÒeÓ after the ÒgÓ. Avoid the stilted expression, ÒOne of us (R. B. G.) thanks . . .Ó  Instead, try ÒR. B. G. thanksÓ. Put sponsor acknowledgments in the unnumbered footnote on the first page.

%%%%%%%%%%%%%%%%%%%%%%%%%%%%%%%%%%%%%%%%%%%%%%%%%%%%%%%%%%%%%%%%%%%%%%%%%%%%%%%%
\bibliographystyle{IEEEtran}
\bibliography{bibliography}

\end{document}